%% file: main.tex
\documentclass[10pt,twocolumn,letterpaper]{article}

\usepackage{iccv}              


\definecolor{iccvblue}{rgb}{0.21,0.49,0.74}
\usepackage[pagebackref,breaklinks,colorlinks,allcolors=iccvblue]{hyperref}

\def\paperID{eOkB8FIISn} 
\def\confName{ICCV}
\def\confYear{2025}

\title{When Less is Enough: Adaptive Token Reduction for Efficient Image Representation}

\author{Eduard Allakhverdov\\
AIRI\\
Moscow, Russia\\
MIPT\\
Dolgoprudny, Russia\\
{\tt\small allakhverdov@2a2i.org}
\and
Elizaveta Goncharova\\
AIRI\\
Moscow, Russia\\
{\tt\small goncharova@airi.net}
\and
Andrey Kuznetsov\\
AIRI\\
Moscow, Russia\\
{\tt\small kuznetsov@airi.net}
}

\begin{document}
\maketitle
\input{sec/0_abstract}    
\input{sec/1_real_intro}
\input{sec/2_related_work}
\input{sec/3_useful_features_en}

\input{sec/4_method_en}
\input{sec/5_experiments_en}

\input{sec/6_discussion_en}

\input{sec/7_conclusion_en}

{
    \small
    \bibliographystyle{ieeenat_fullname}
    \bibliography{main}
}

\end{document}

%% file: sec/0_abstract.tex
\begin{abstract}
Vision encoders typically generate a large number of visual tokens, providing information-rich representations but significantly increasing computational demands. This raises the question of whether all generated tokens are equally valuable or if some of them can be discarded to reduce computational costs without compromising quality. In this paper, we introduce a new method for determining feature utility based on the idea that less valuable features can be reconstructed from more valuable ones. We implement this concept by integrating an autoencoder with a Gumbel-Softmax selection mechanism, that allows identifying and retaining only the most informative visual tokens. To validate our approach, we compared the performance of the LLaVA-NeXT model, using features selected by our method with randomly selected features. We found that on OCR-based tasks, more than 50\% of the visual context can be removed with minimal performance loss, whereas randomly discarding the same proportion of features significantly affects the model capabilities. Furthermore, in general-domain tasks, even randomly retaining only 30\% of tokens achieves performance comparable to using the full set of visual tokens. Our results highlight a promising direction towards adaptive and efficient multimodal pruning that facilitates scalable and low-overhead inference without compromising performance.
\end{abstract}

\begin{figure}[t]
  \centering
   \includegraphics[width=0.8\linewidth]{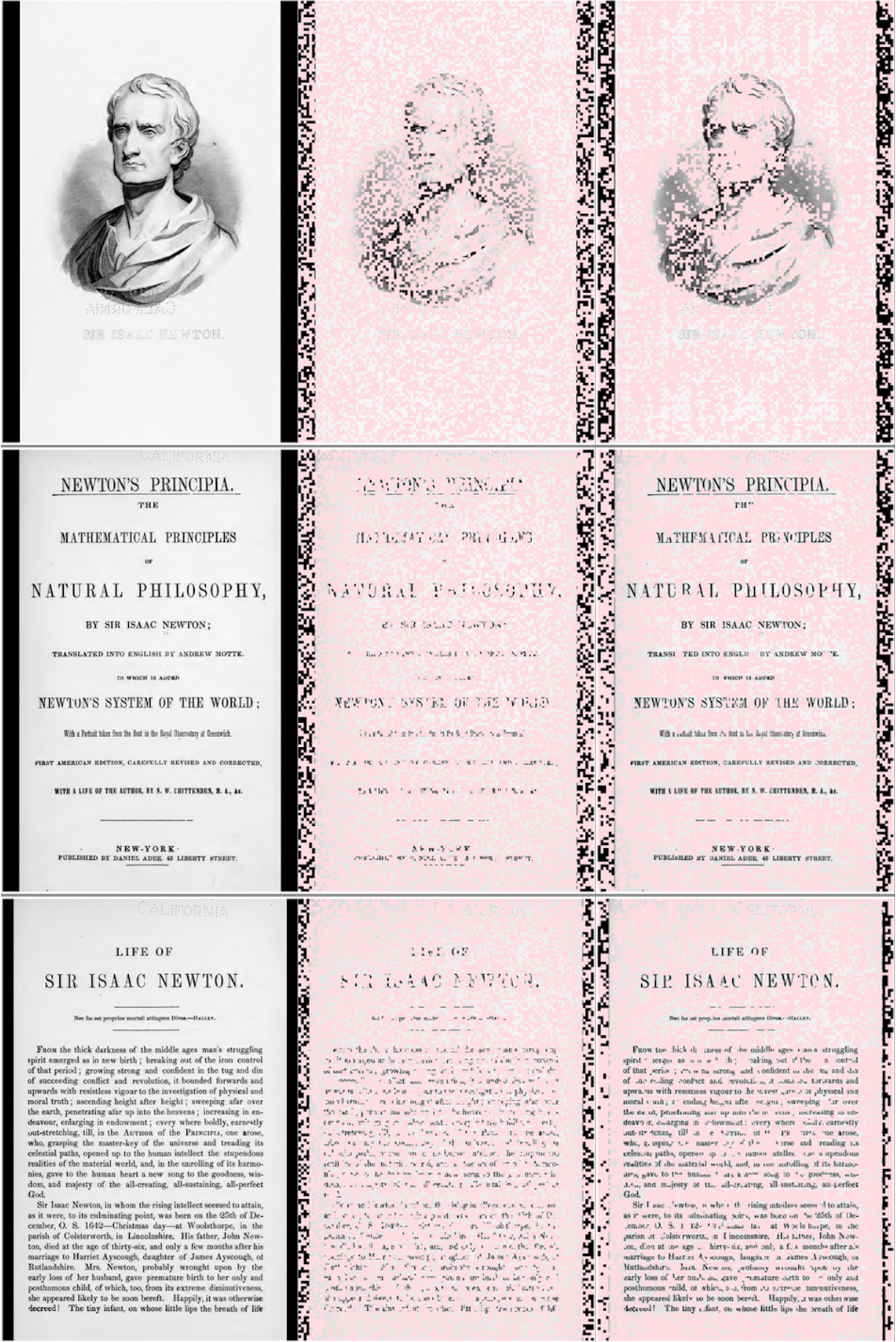}
   \caption{Comparison of feature selection methods on Newton's Principia text: original image (left), random feature selection retaining 40\% of tokens (middle), and our proposed feature selector retaining 40\% of tokens (right).}
   \label{fig:newton}
\end{figure}

%% file: sec/1_real_intro.tex
\section{Introduction}

In recent years, vision encoders have become important components for various downstream tasks, providing universal representation of visual features. These encoders are trained to effectively compress raw pixel information into latent embeddings. Depending on their training objectives, vision encoders can encapsulate different types of information in their hidden states. However, it is widely recognized that many of these encoded features contain redundant or irrelevant information for downstream tasks~\cite{raghu2022visiontransformerslikeconvolutional, naseer2021intriguingpropertiesvisiontransformers, tong2024cambrian1fullyopenvisioncentric}. Therefore, reducing the number of output features produced by vision encoders is an important and challenging task --- especially now, as encoders increasingly serve as fundamental mechanisms for visual understanding in multimodal models~\cite{li2024llavanext, chen2024internvlscalingvisionfoundation, tong2024cambrian1fullyopenvisioncentric}.

Multimodal models that process visual inputs typically condition on outputs of a Vision Transformer (ViT) \cite{dosovitskiy2021imageworth16x16words}, appending a long vision-derived prefix to the input of a Large Language Model (LLM) via a projection layer. Although this method gives promising results, handling large context length (especially when processing high-resolution images) remains a significant challenge. Moreover, previous studies have observed that not all ViT outputs equally contribute to downstream task performance~\cite{devoto2024adaptivelayerselection}; many tokens can be redundant, noisy, or simply irrelevant~\cite{yang2024denoisingvisiontransformers}. Therefore, selectively identifying and retaining only the most informative features can significantly decrease the number of tokens while maintaining model performance.

To address this issue, we propose a novel method to select the most informative visual features from the encoder output using an autoencoder-based approach implemented with Gumbel-Softmax sampling. Our method identifies features that are essential to preserve crucial visual information, allowing us to accurately reconstruct the original feature set. We show that this training procedure not only efficiently identifies valuable features, but also provides highly interpretable results. Furthermore, we illustrate how our approach can be seamlessly integrated during inference in multimodal models, significantly reducing the visual prefix length without compromising performance.

In experiments conducted with the LLaVA-NeXT \cite{li2024llavanext} and LLaVA-OneVision~\cite{li2024llavaonevisioneasyvisualtask} models, we demonstrate that features selected using the the proposed approach contain essential information for the model to provide the correct answer to most of the analyzed tasks. Notably, our method reduces visual context length by up to 50\% with minimal performance degradation in most benchmarks --- and achieves reductions of up to 90\% in certain tasks.

The contributions of our paper can be summarized as follows:
\begin{itemize}
    \item We propose a novel, interpretable method for selecting the most informative features from vision encoders.
    \item We demonstrate how our approach serves as an effective in-situ feature reduction method for existing multimodal models without requiring further fine-tuning.
    \item We empirically confirm that retaining as little as 50\% of the original visual features can be sufficient to maintain near-baseline performance on multiple multimodal benchmarks.
\end{itemize}

%% file: sec/2_related_work.tex
\section{Related Work}

Several approaches to reduce redundancy in token-based visual representations have been proposed in recent literature. We will discuss the most relevant ones below.

\subsection{Token Pruning}

Token pruning is a widely used method to remove irrelevant or noisy features from Vision Transformers while preserving the most valuable information. Many token pruning methods rely on attention scores to guide feature selection~\cite{tang2023}. To further improve versatility, some approaches propose to retain not only the most relevant but also the most diverse set of tokens, thereby preserving a richer representation of the image~\cite{long2023}. Maintaining high quality embeddings is critical, especially in detection and segmentation tasks, as they require detailed and precise image representation~\cite{Liu_2024_WACV}.

Token pruning methods are often tailored to specific vision tasks. For example, \citet{Kinfu2023EfficientVT} proposes three specialized pruning methods for human pose estimation. Two of these methods use a lightweight pose estimation network to guide patch selection, while the third uses learnable joint tokens to explicitly ensure that the selected patches retain key body joint information. The overall goal is to preserve the most informative embeddings needed for accurate pose estimation.

\subsection{Token Generation and Merging}

Another line of research involves generating or merging tokens to obtain a compact yet informative subset from the original set of tokens. TokenLearner~\cite{Ryoo2021} merges hidden representations within Vision Transformers to produce a concise output containing only eight tokens that encapsulate internal functionality. Similarly, Token Merger~\cite{Feng2023} introduces the concept of ``meta-tokens,'' adaptively merging similar tokens to retain essential information. In~\cite{lee2024}, learnable decoupled embeddings are used, allowing end-to-end token to be merged without relying solely on similarity measures. In addition, Resizable-ViT~\cite{Zhou2023longshort} introduces a flexible module that learns token-length labels, identifying the most informative tokens for detailed analysis.

Some models adopt a fixed, stage-wise downsampling strategy. For example, the Pyramid Vision Transformer (PVT)~\cite{Wang2021pvt} divides the input image into increasingly coarser token maps at deeper layers, mimicking the pyramidal structure of CNNs. Unlike adaptive token merging~\cite{Feng2023}, PVT statically reduces sequence length at each stage, thereby reducing computational cost for high-resolution inputs. Although this hierarchical design has proven effective for tasks such as detection and segmentation, it does not allow dynamic pruning or merging of tokens based on their content.

In addition, \citet{li2023patchneed} explores the broader implications of tokenization in Vision Transformers, proposing convolutional patterns to enhance visual representation quality.

While these token pruning and merging methods show promising results on traditional computer vision tasks (classification, segmentation, detection), they usually lack versatility. Most existing approaches cannot be directly transferred to multimodal models, which must leverage as much relevant visual information as possible in conjunction with textual inputs.

\subsection{Vision Context Reduction in Multimodal Models}

Reducing the size of the visual context is particularly important in multimodal models, where visual tokens significantly increase the context length required by the LLM. However, visual tokens vary in their relevance depending on the query. Various strategies have been developed to address this problem. For example, interpolation-based methods reduce the number of visual tokens by attempting to preserve critical visual information from the entire feature subset (e.g., LLaVA-OneVision~\cite{li2024llavaonevisioneasyvisualtask}). InternVL series of models~\cite{chen2025expandingperformanceboundariesopensource} use a pixel-unshuffle operation typically applied to high-resolution images. Other popular approaches use trainable modules that either compress visual tokens or introduce learnable queries to extract important visual information, such as Perceiver IO~\cite{jaegle2022perceiveriogeneralarchitecture} and BLIP-2~\cite{li2023blip2bootstrappinglanguageimagepretraining}.

In our work, we propose a task-agnostic method to directly select the most informative visual features from the encoder outputs without fine-tuning or dependence on textual input. Therefore, our approach can be seamlessly applied to both purely visual tasks and multimodal scenarios.

%% file: sec/3_useful_features_en.tex
\section{Useful Feature Selection}
\label{sec:intuition}

The Transformer architecture has been successfully used as a backbone for vision encoders \cite{dosovitskiy2021imageworth16x16words}, providing hidden representations suitable for a wide range of vision tasks. However, due to the inherent design of the self-attention mechanism in Transformers, neighboring tokens naturally contain information about each other. Consequently, we assume that information may be duplicated redundantly in different regions of the output feature tensor. In particular, some visual representations could potentially be composed entirely of information already present in other tokens. If such redundant representations exist, they can be identified and removed without causing significant performance degradation in vision-related tasks.

This hypothesis naturally raises two critical questions: how can one quantitatively measure whether one set of features contains more information than another, and how can one select the optimal subset of features?

\subsection{Feature Subset Comparison}
\label{subsec:who_is_better}

For any image $I$,  the corresponding feature set $F$ has dimensions $(L, C)$. As mentioned earlier, tokens identified for potential exclusion have the characteristic property that they can be reconstructed from the remaining visual tokens in the set. Thus, if there exists an optimal reconstruction function $R$, which takes a pruned subset of features as input $F^{pr}$ (where the superscript $pr$ denotes \textit{pruned}) with dimensions $(L^{pr}, C)$ and returns a reconstructed set $F^{rec}$ with dimensions $(L, C)$, and if a proximity function $dist$ is defined between two tensors, then one subset is considered superior to the other if it allows for a more accurate reconstruction of the discarded visual tokens.

Formally, subset $F^{pr}_1$ is superior to subset $F^{pr}_2$ if:

\begin{equation}
  dist\left(R(F^{pr}_1), F\right) < dist\left(R(F^{pr}_2), F\right).
  \label{eq:important}
\end{equation}

\subsection{How to Select the Optimal Set?}
\label{subsec:min_problem}

To select the most informative features, we aim to find a function \(S\), referred to as the \emph{optimal selector}, which takes \(F\) as input and returns a pruned subset \(F^{pr}\). We train this selector in a way similar to the autoencoder:

\begin{equation}
 \min_{\theta, \psi}\ \text{dist}\Big(R_{\psi}(S_{\theta}(F)),\, F\Big) + L^{pr}.
 \label{eq:important}
\end{equation}

In this formulation, the first term provides a high-quality reconstruction of the original feature set from the pruned subset. The additional term \(L^{pr}\) penalizes the selector \(S_{\theta}\) if it trivially selects all tokens (acting as an identity function), thereby encouraging a more concise but informative subset.

%% file: sec/4_method_en.tex
\section{Method}
\label{sec:method}

\subsection{Implementation Details}
\label{sec:implementation}

In this section, we present the implementation details of the approach described in \cref{sec:intuition}, which consists of two main components: \emph{Feature Selector} and \emph{Feature Reconstructor}.

\subsubsection{Feature Selector Architecture}
\label{subsec:S}

\begin{figure}[t]
  \centering
  \includegraphics[width=0.8\linewidth]{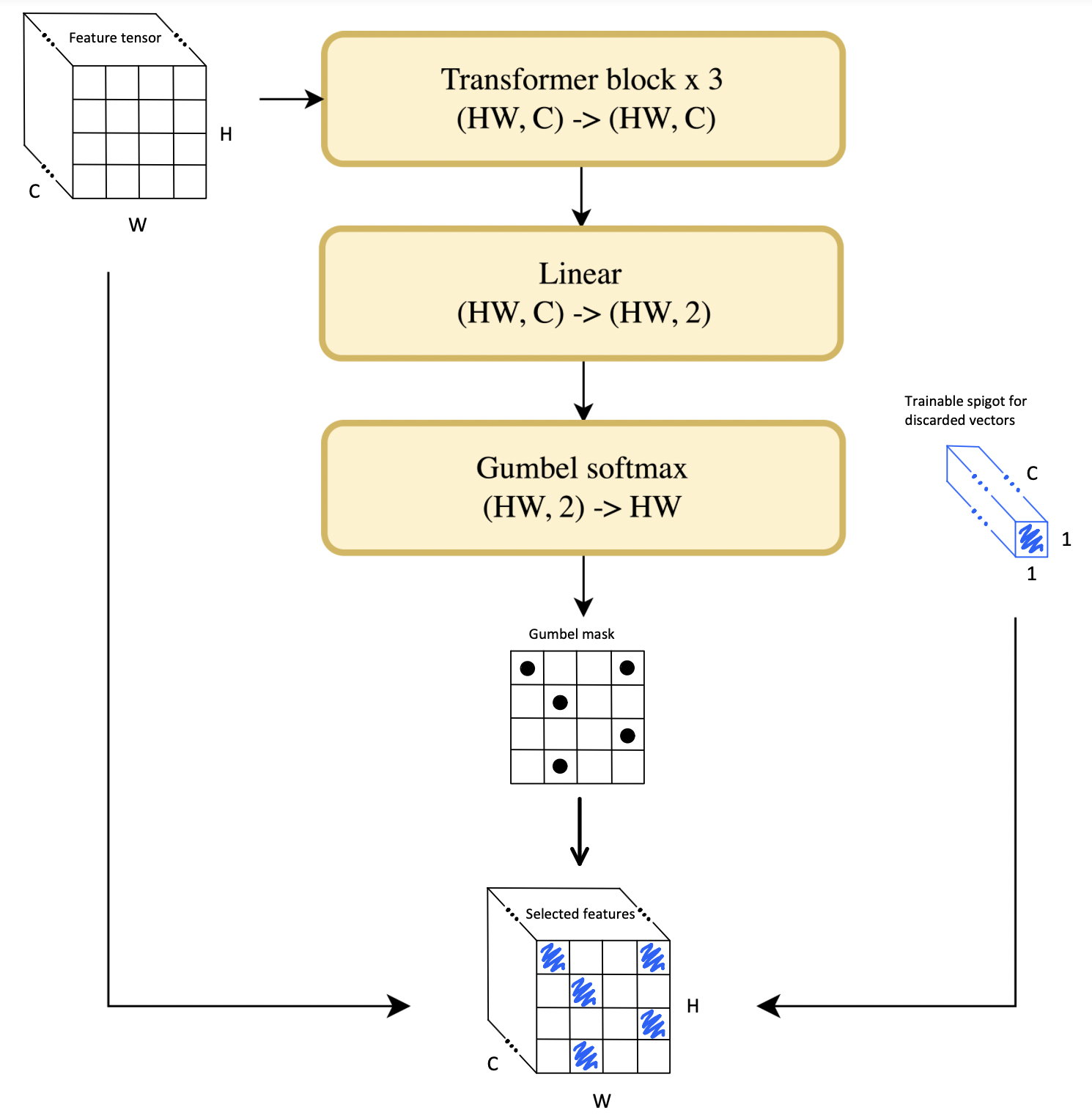}
  \caption{Illustration of the Feature Selector in training mode. It uses three Transformer layers and a Gumbel-Softmax head to generate a binary mask where zeros mark tokens for removal and ones for retention. During training, the masked embeddings are replaced by a shared learnable embedding. During inference, the masked embeddings are discarded, while the retained ones are used for downstream tasks, such as image representations in Vision-Language models.}
  \label{fig:S}
\end{figure}

Feature Selector \(S\) consists of three Transformer layers and a Gumbel-Softmax-based \cite{jang2017gumbelsoftmax} head. The head creates a binary mask, as shown in \cref{fig:S}, where zeros indicate visual tokens to be removed and ones indicate tokens to be retained.

During training, feature embeddings corresponding to zeros in the binary mask are replaced by a shared learnable embedding $E_{masked}$ (this embedding will be reconstructed later by the component described in \cref{subsec:R}). During inference, embeddings corresponding to zeros are simply discarded, while those corresponding to ones are kept for downstream task. For example, they can be used as image representations in Vision-Language models, as shown in our experiments in \cref{sec:experiments}.

For more flexibility during inference, one can choose to use logits from the linear layer instead of a hard binary mask. Based on these logits, the user can select a fixed number of the most informative features. This is exactly the approach that is used in our experiments, which we describe in \cref{sec:experiments}.

\subsection{Feature Reconstructor Architecture}
\label{subsec:R}

\begin{figure}[t]
  \centering
  \includegraphics[width=0.8\linewidth]{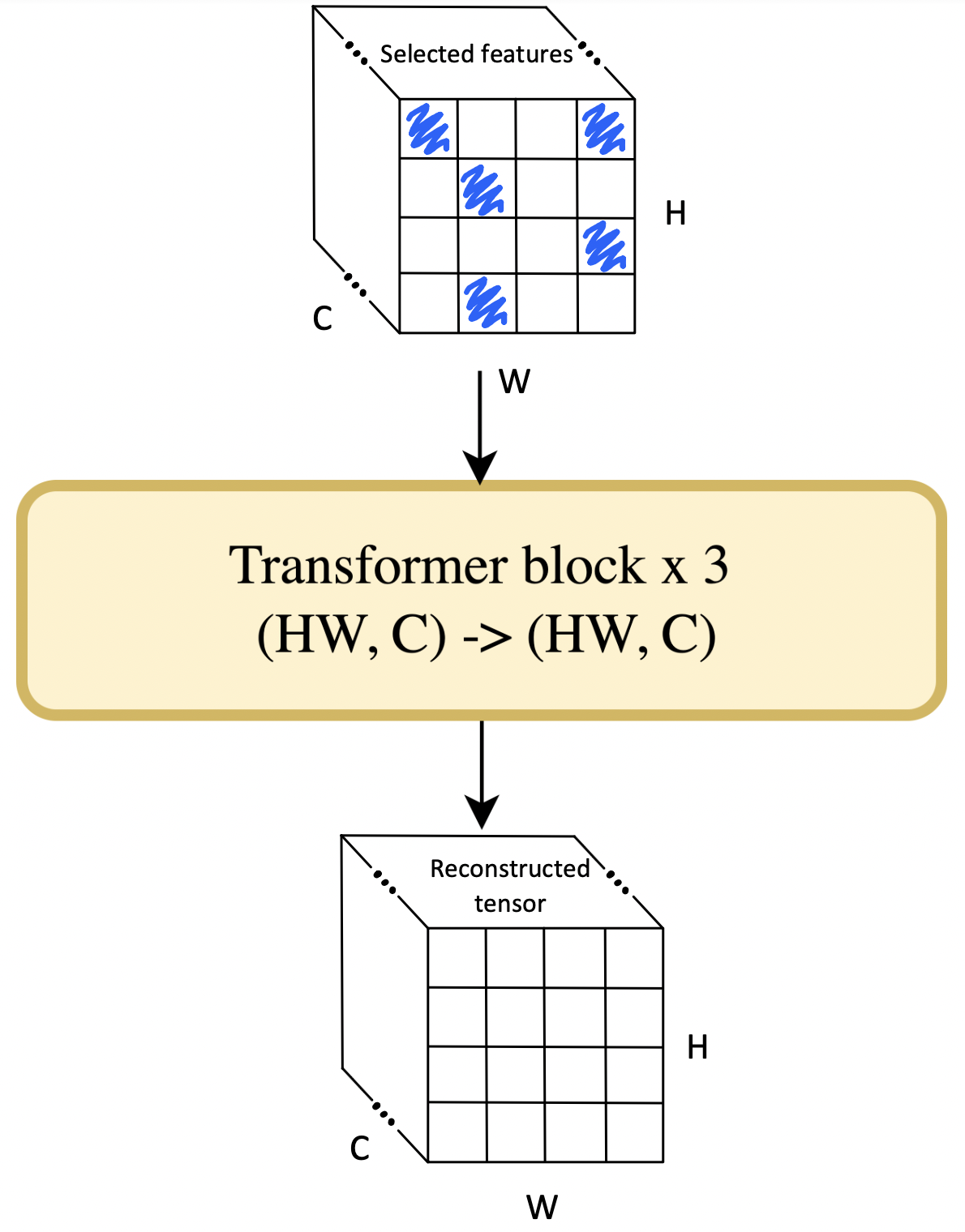}
  \caption{Illustration of Feature Reconstructor's functionality. Its primary objective is to restore the tokens that were replaced with a learned representation.}
  \label{fig:R}
\end{figure}

Feature Reconstructor \(R\) also consists of three Transformer layers. Its primary objective is to restore the tokens that were replaced with a learned embedding $E_{masked}$, as illustrated in \cref{fig:R}.

\subsection{Loss Function}
\label{subsec:loss}

As described in \cref{subsec:min_problem}, the optimization objective is formulated as the sum of two terms: (1) a reconstruction loss and (2) a regularization term that aims to minimize the amount of information required for reconstruction. In principle, we would expect the reconstruction loss to approach zero while the regularization term converges to the fraction of useful visual tokens. However, in practice we did not observe the expected behavior. We found that the optimizer is more likely to converge to a local minimum where the regularization term drops to zero, thereby avoiding the token utilization penalty.

To resolve this issue, we modify the regularization term as follows.

\begin{equation}
    L^{pr} \;\rightarrow\; \max(L^{pr},\, p),
    \label{eq:max_reg}
\end{equation}
where \(p\) is a predetermined proportion of useful features.

In other words, whenever \(L^{pr}\) falls below \(p\), the regularization penalty is effectively disabled, allowing for further optimization of the reconstruction loss. Empirically, we observe that \(L^{pr}\) initially decreases to \(p\) and then fluctuates around it during the remainder of training, while the reconstruction loss continues to decrease. 

\subsection{Training}
\label{subsec:training}

As shown in \cref{fig:S,fig:R}, our approach is similar to VQ-VAE \cite{oord2018vq-vae}, except that we use a set of input features instead of a learned dictionary, and the latent representation may vary in size.

We train both \(R_{\psi}\) and \(S_{\theta}\) following the framework introduced in
\cref{subsec:min_problem}. Specifically, we choose the \(l_{2}\) norm for the distance function \(\mathrm{dist}\) and compute \(L^{pr}\) using the mask generated by \(S_{\theta}\). 

\paragraph{Feature Selector.} 

The feature selector \(S_{\theta}\) processes the original feature tensor \(F\) and outputs a subset of selected features \(F^{pr}\) along with a binary mask \(M\), referred to as the ``Gumbel mask,'' as illustrated in \cref{fig:S}. Formally, this can be expressed as:
\begin{equation}
  F^{pr},\; M \;=\; S_{\theta}(F),
  \label{eq:selector}
\end{equation}
where the mask \(M\) specifies which spatial locations of the input tensor \(F\) are retained (marked as ones) and which are discarded (marked as zeros). The output \(F^{pr}\), labeled as ``Selected features'' in \cref{fig:S}, is formed by replacing the discarded feature vectors with a shared learnable representation (shown as blue hatched vectors).

\paragraph{Feature Reconstructor.}

The reconstructor is defined by:
\begin{equation}
  F^{rec} \;=\; R_{\psi}(F^{pr}),
  \label{eq:reconstructor}
\end{equation}
with \(F^{rec}\) denoting the ``Reconstructed tensor'' shown in \cref{fig:R}.

\paragraph{Regularization Term.}
Regularization term is computed directly from the mask:
\begin{equation}
    L^{pr} \;=\; \sum_{h=0, w=0}^{H, W} \frac{M_{h, w}}{HW}.
    \label{eq:regularization}
\end{equation}

\paragraph{Overall Objective.}

Incorporating the modified regularization from \cref{subsec:loss}, the overall optimization problem can be defined as follows:
\begin{equation}
  \min_{\theta, \psi}\;\Big\|F^{rec} - F\Big\|_{2} \;+\; \max\!\big(L^{pr},\, p\big).
\label{eq:minimization_problem}
\end{equation}

All components are fully differentiable, and we optimize them using gradient descent.

\subsection{Dataset}

For our training dataset, we sampled 100K images from the COCO dataset \cite{lin2015microsoftcococommonobjects}. Each image was pre-processed with a specific vision encoder for which the selector was trained. The resulting feature representations were used as training data.


%% file: sec/5_experiments_en.tex
\section{Experiments}
\label{sec:experiments}

To evaluate the performance of the proposed feature selector, we integrated it with vision encoders used as backbones in two multimodal models: LLaVA-NeXT (visual encoder of CLIP~\cite{radford2021}) and LLaVA-OneVision (visual encoder of SigLIP~\cite{zhai2023siglip}). Our primary goal was to investigate whether the selector described in \cref{subsec:S} can be applied directly (i.e., without further fine-tuning) to existing multimodal architectures, and to determine whether selecting informative features positively impacts model inference. To this end, we evaluated both models augmented with our trained feature selector under different pruning factors, comparing their performance to a baseline using random feature selection with the same pruning factors across multiple multimodal benchmarks.

\subsection{LLaVA-NeXT: Our Selector vs.\ Random Selector}
\label{subseq:llava-next-res}

\begin{figure}[t]
  \centering
  \includegraphics[width=0.8\linewidth]{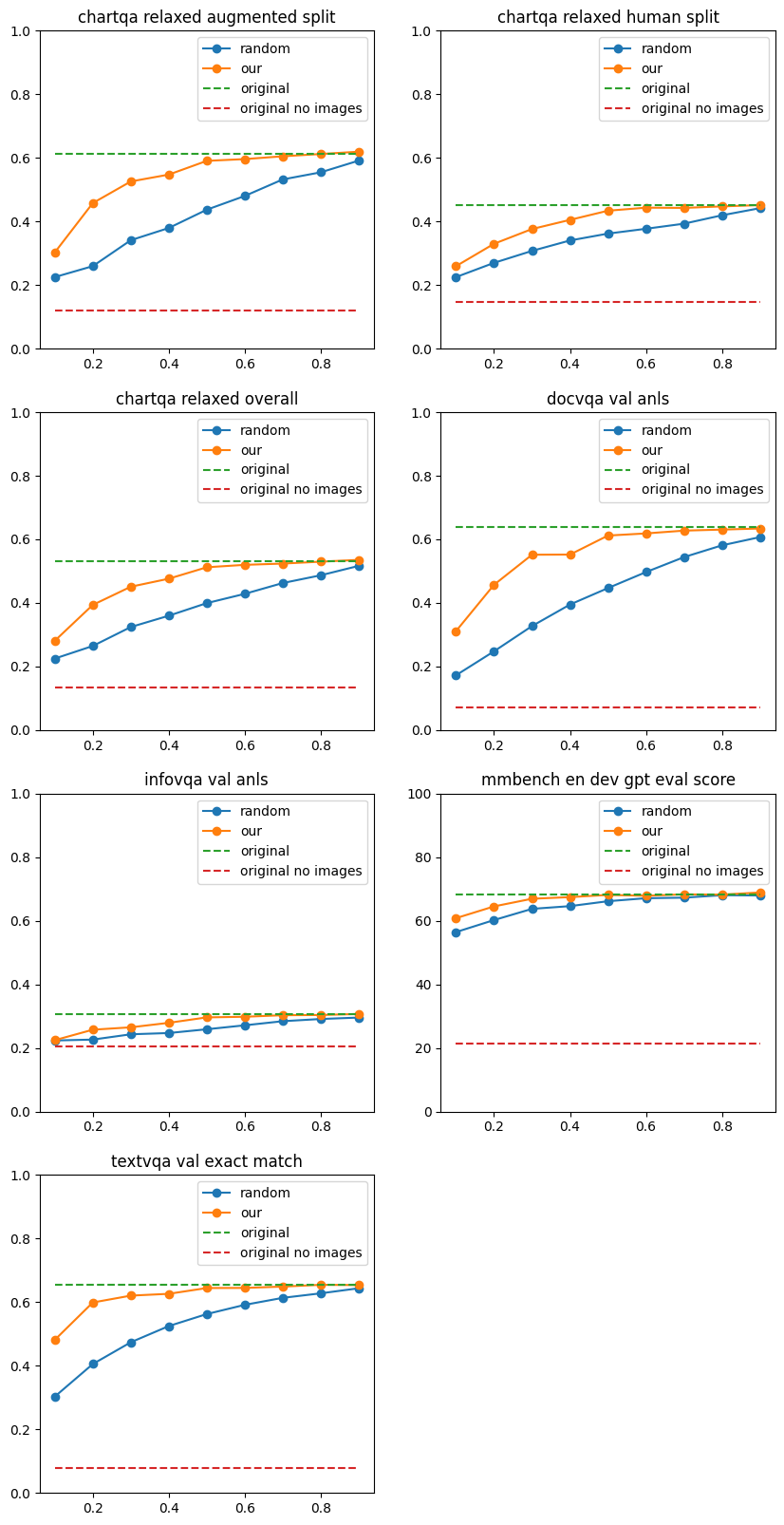}
  \caption{Comparison of LLaVA-NeXT performance with our selector (orange) and random selector (blue) on text-based benchmarks. The green dashed line represents the baseline performance using all features. The red dashed line represents the model's performance without image input.}
  \label{fig:llava-next-text}
\end{figure}

\begin{figure}[t]
  \centering
  \includegraphics[width=0.9\linewidth]{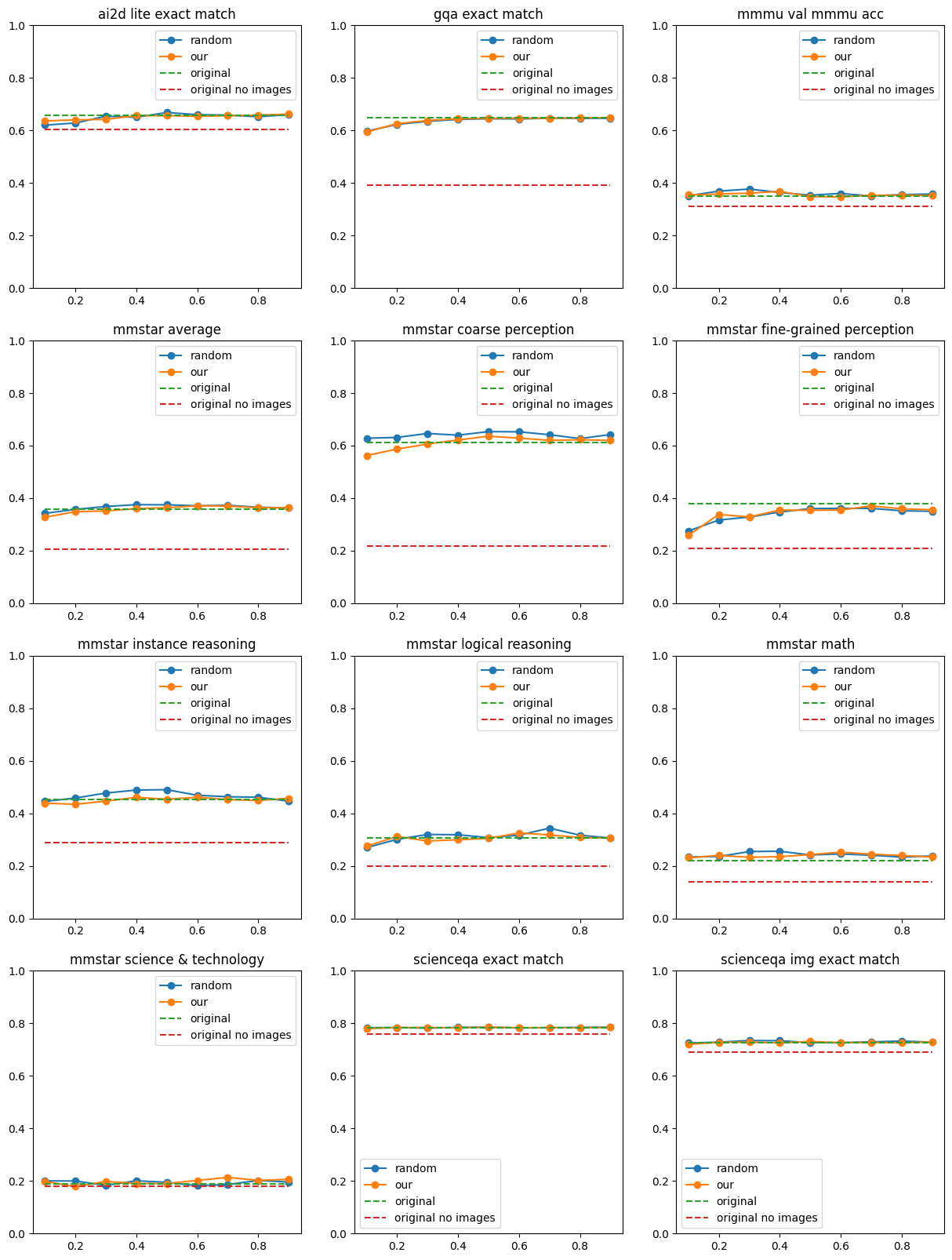}
  \caption{Comparison of LLaVA-NeXT performance with our selector (orange) and random selector (blue) on non-text benchmarks. The green dashed line represents the baseline performance using all features. The red dashed line represents the model's performance without image input.}
  \label{fig:llava-next-no-text}
\end{figure}

Figures~\ref{fig:llava-next-text} and~\ref{fig:llava-next-no-text} present the experimental results obtained using the LLaVA-NeXT model in various multimodal benchmarks. For clarity, the results are organized into two groups. The first group (\cref{fig:llava-next-text}) contains OCR-like benchmarks involving large images, including DocVQA~\cite{mathew2021docvqa}, ChartQA~\cite{masry2022chartqa}, InfoVQA~\cite{mathew2021infographicvqa}, TextVQA~\cite{singh2019vqamodelsread}, and MMBench~\cite{liu2024mmbench}. While MMBench does not consist entirely of OCR-based tasks, it includes complex tasks that require image comprehension. The second group (\cref{fig:llava-next-no-text}) includes the other benchmarks that evaluate the ability of models to solve various academic tasks and perform scene understanding (AI2D~\cite{Hiippala_2020}, GQA~\cite{hudson2019gqa}, MMMU~\cite{yue2024mmmu}, MMStar~\cite{chen2024mmstar}, and ScienceQA~\cite{lu2022scienceqa}).

In each figure, the horizontal axis represents the proportion of features retained (i.e., the parameter \(p\) defined in~\cref{subsec:loss}), while the vertical axis indicates the evaluation metric obtained on each benchmark. There we compare the performance of the original model using all available features (green dashed line) with the model using features selected by our trained selector (orange line) and randomly selected features (blue line). We also measure the original model without showing images (red dashed line) to understand how important it is to see the image when solving this benchmark.

From the OCR-like benchmark results (\cref{fig:llava-next-text}), two primary conclusions can be drawn:
\begin{enumerate}
    \item The trained selector significantly outperforms random selection across all tested retention ratios.
    \item \textcolor{PineGreen}{\textbf{Up to 50\% of visual features can be discarded using the trained selector with only negligible performance degradation.}}
\end{enumerate}

From the remaining benchmarks (see \cref{fig:llava-next-no-text}), we observe two additional outcomes:
\begin{enumerate}
    \item The quality of the model's responses remains largely unaffected by the number of visual features retained and closely matches the baseline with all features.
    \item \textcolor{BrickRed}{\textbf{In these tasks, the choice of feature selection method (trained vs.\ random) has minimal impact on the quality of the response.}}
\end{enumerate}

\subsection{LLaVA-OneVision: Our Selector vs.\ Random Selector}
\label{subsec:llava-onevision-results}

The LLaVA-OneVision model includes a built-in compression mechanism, which reduces the size of the visual feature tensor through interpolation. In the LLaVA-OneVision implementation, interpolation compression is activated from a predefined reference size, and the compression ratio varies depending on the input image dimensions. For example, larger documents typically result in compression ratios of 1.7 to 2.

Since our main goal is to demonstrate that our proposed selector effectively identifies informative visual features, we disabled the default compression mechanism in LLaVA-OneVision instead of combining two distinct compression methods.

During preliminary experiments, we noticed that if the visual context exceeds the internally fixed context length of the model, the performance tends to degrade. Therefore, we limited our experiments to the range \(p \in [0.1,\, 0.2,\, 0.3,\, 0.4,\, 0.5,\, 0.6 \approx \frac{1}{1.7}]\), ensuring that the length of the retained visual features does not exceed approximately \(\frac{1}{1.7}\) the original input length.

\begin{figure}[t]
  \centering
  \includegraphics[width=0.8\linewidth]{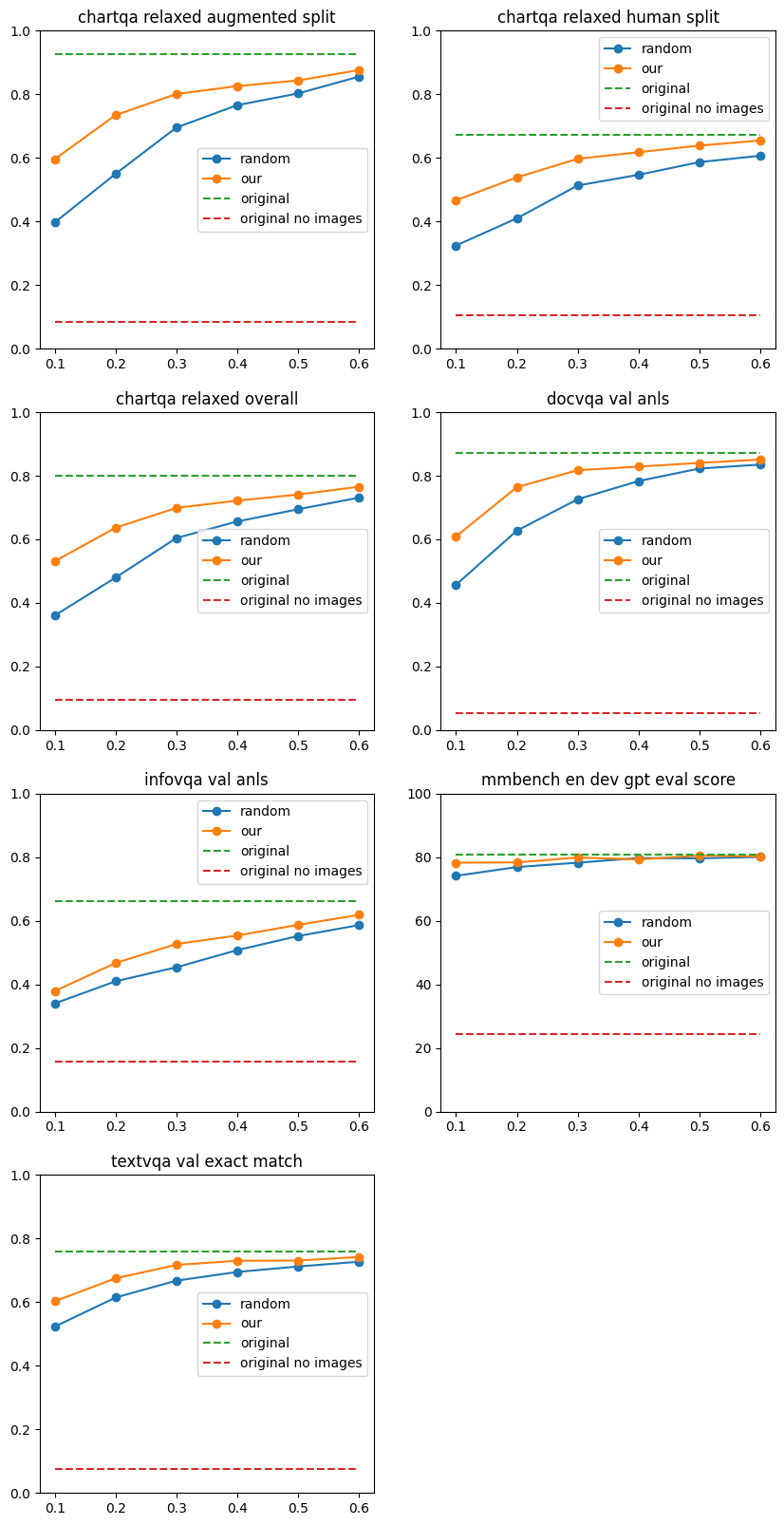}
  \caption{Comparison of LLaVA-OneVision performance with our trained selector (orange) and random selection (blue) on text-based (OCR-like) benchmarks. The green dashed line represents the baseline performance with all features retained. The red dashed line represents the model's performance without image input.}
  \label{fig:ov-text}
\end{figure}

\begin{figure}[t]
  \centering
  \includegraphics[width=0.9\linewidth]{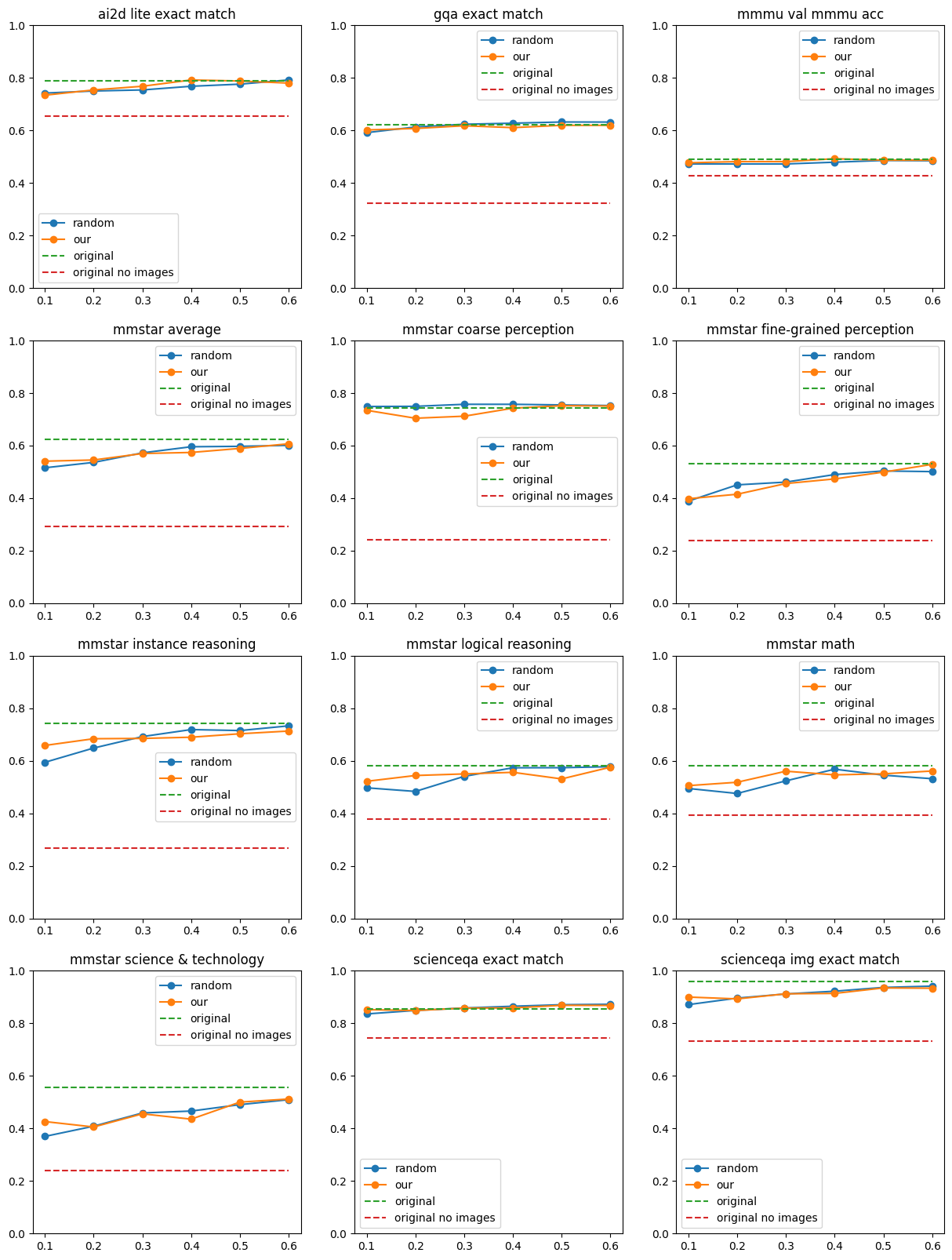}
  \caption{Comparison of LLaVA-OneVision performance with our trained selector (orange) and random selection (blue) on non-text benchmarks. The green dashed line represents the baseline performance with all features retained. The red dashed line represents the model's performance without image input.}
  \label{fig:ov-no-text}
\end{figure}

Similar to the LLaVA-NeXT evaluations, the results for LLaVA-OneVision are divided into two categories, as depicted in \cref{fig:ov-text,fig:ov-no-text}.

For OCR-based benchmarks involving high-resolution images (\cref{fig:ov-text}), our trained selector consistently outperforms random selection, proving its ability to effectively preserve informative visual features.

In contrast, for non-OCR benchmarks (\cref{fig:ov-no-text}), the results are very similar to those observed previously with LLaVA-NeXT (\cref{fig:llava-next-no-text}), indicating that the performance is largely unaffected by the choice of selection method in these scenarios.

%% file: sec/6_discussion_en.tex
\section{Discussion}
\label{sec:discussion}

Given the experimental results presented in~\cref{sec:experiments}, an important question arises: why does a trained feature selector significantly improve performance on OCR-like tasks involving high-resolution images, but not consistently on other tasks? The following analysis attempts to explain this discrepancy.

The strongest difference in performance depending on the percentage of visible tokens is observed on ChartQA, DocVQA, and TextVQA, InfoVQA benchmarks. These tasks have a common characteristic: to provide the correct answer, the model must identify the detailed visual information in the image, without extensive logical reasoning afterwards. In other words, these tasks rely heavily on visual perception and minimally on subsequent cognitive reasoning, making the advantage of selectively retaining informative features particularly pronounced.

In contrast, MMBench also requires careful perception of visual details, but solving them additionally requires logical reasoning. As a result, the overall performance depends not only on the quality of visual input, but also significantly on the reasoning capability of the language model, thereby reducing the relative advantage provided by selective feature sampling. However, we still observe that our feature selector outperforms the random one. 

For the above benchmarks, we see a consistent pattern where metrics improve substantially as the percentage of visible features increases from 10\% to approximately 50\%. Beyond this threshold, performance gains become increasingly insignificant.

The tasks represented by the MMMU and MMStar datasets (in particular, the logical reasoning, mathematical reasoning, and science \& technology categories) have the highest reasoning complexity. Here, the language model's reasoning abilities play a predominant role, effectively overshadowing the impact of feature selection strategies.

Finally, tasks from AI2D, GQA, ScienceQA, and certain categories in MMStar (e.g., coarse perception, fine-grained perception, and instance reasoning) rely on a general understanding of the scene rather than isolated visual details. For these benchmarks, comprehensive feature coverage across the image proves more important than the selective retention of individually informative features. Consequently, random selection can perform comparable to or even outperform intelligent sampling in some cases due to more uniform coverage of the entire scene (e.g., in MMStar's coarse perception and instance reasoning tasks).

It is also interesting to note the performance of models inferenced without a visible image, and the different percentages of sampled image features. As we can see for the MMStar benchmark, there is a gap between the no image and the entire image observed by the model, accounting for 10-20\% performance improvement on average. Retaining only 10\% of the features maintains performance comparable to using the entire image. For the MMMU benchmark, the behavior is noticeably different. We can observe (see~\cref{fig:llava-next-no-text,fig:ov-no-text}) that there is no significant difference between no image and the entire image. Performance on this benchmark depends primarily on the quality of the language model rather than on the ability to perceive visual information.

\begin{figure}[t]
  \centering
   \includegraphics[width=0.9\linewidth]{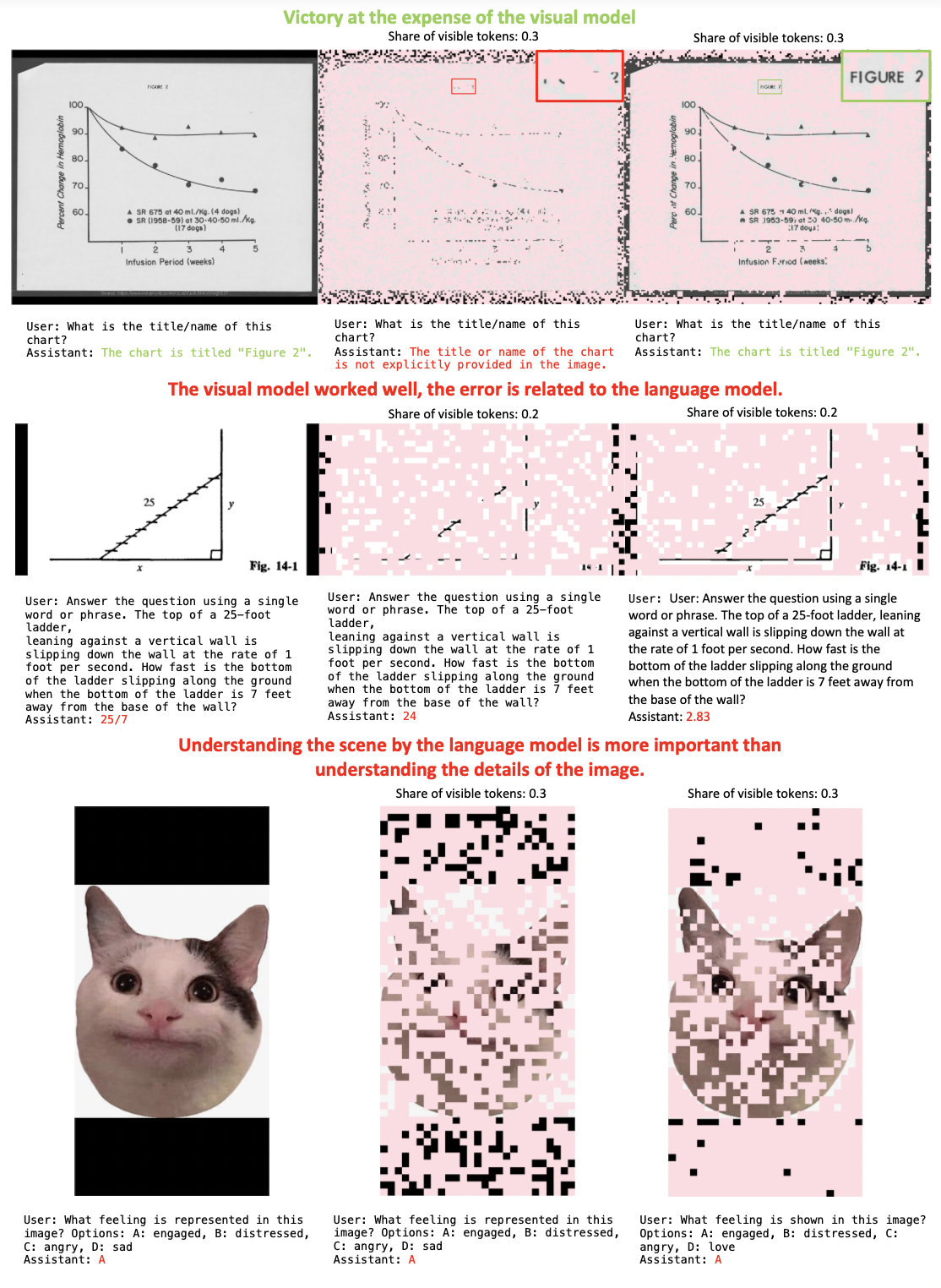}
   \caption{Images from three benchmarks illustrating cases where the vision-language model gives correct answers or makes errors. The first column shows the model's responses using the full visual context, the second column uses a randomly selected set of features, and the third column uses the features selected by our selector. (1) DocVQA: to answer the question selecting the correct features is crucial. (2) MMMU (math): to answer this question, both visual understanding and logical reasoning are important, but the model fails to reason correctly. (3) MMstar: the image details are less important, and the language model plays a dominant role.}
   \label{fig:3_examples}
\end{figure}

In \cref{fig:3_examples}, we provide examples of the images from three different benchmarks, illustrating the sampling mechanism and the types of questions for which the vision-language model either produces correct answers or makes errors.

%% file: sec/7_conclusion_en.tex
\section{Conclusion}
\label{sec:conclusion}

In this paper, we proposed a novel method to select informative features from visual encoders. Our method is based on trainable VAE module with the Gumbel-Softmax built-in the ViT model. The proposed method allows to reduce the number of output vision tokens, while maximally preserving the most informative ones. To validate the effectiveness of our approach, we demonstrated how the proposed module can serve as an efficient feature reduction strategy for multimodal models. Experimental results show that up to 50\% of the visual context can be removed with minimal performance degradation for the LLaVA-NeXT and LLaVA-OneVision models. 

Furthermore, our feature selection method demonstrates high interpretability, as illustrated by qualitative examples given in~\cref{fig:newton}, where the selected features clearly correspond to specific objects and shapes in the images.

However, we acknowledge certain limitations of the proposed method, including limited compatibility with interpolation-based feature compression methods that are widely used in modern multimodal models. To address this limitation, we plan to study joint fine-tuning of the feature selector and the language model as future work. Such fine-tuning can potentially mitigate compatibility issues that arise when combining different compression strategies (e.g., interpolation-based and Gumbel-based selections).

We believe that these results open promising new directions for research aimed at extracting informative representations from visual contexts. Potential applications of this approach include faster inference, reduced memory consumption, and improved resilience to noisy visual inputs.